%% file: main.tex
\documentclass{article}
\pdfoutput=1
\PassOptionsToPackage{numbers}{natbib}
\usepackage[final]{style}

\usepackage[utf8]{inputenc} % allow utf-8 input
\usepackage[T1]{fontenc}    % use 8-bit T1 fonts
\usepackage{hyperref}       % hyperlinks
\usepackage{url}            % simple URL typesetting
\usepackage{booktabs}       % professional-quality tables
\usepackage{amsfonts}       % blackboard math symbols
\usepackage{nicefrac}       % compact symbols for 1/2, etc.
\usepackage{microtype}      % microtypography
\usepackage{comment}
\usepackage{multirow}
\usepackage[labelfont=bf]{caption}
\usepackage{capt-of}
\usepackage{graphicx}
\usepackage{placeins}
\usepackage{changepage}
\usepackage{caption} 
\usepackage{titlesec}

\usepackage{floatrow}
\newfloatcommand{capbtabbox}{table}[][\FBwidth]

\newcommand{\para}{\textbf}

\newcommand\figcaption[2]{\captionof{figure}{\textsl{\small\textbf{#1} #2}}}
\newcommand\tablecaption[2]{\captionof{table}{\textsl{\small\textbf{#1.} #2}}}

\titlespacing\section{0pt}{8pt plus 4pt minus 2pt}{0pt plus 2pt minus 2pt}
\titlespacing\subsection{0pt}{8pt plus 4pt minus 2pt}{0pt plus 2pt minus 2pt}

\title{The built environment and induced transport CO\textsubscript{2} emissions: A double machine learning approach to account for residential self-selection}
\author{%
    Florian Nachtigall \\
    Technical University of Berlin\\
    Berlin, Germany \\
    \texttt{nachtigall@tu-berlin.de} \\
  \And
    Felix Wagner \\
    Technical University of Berlin\\
    Berlin, Germany \\
  \And
    Peter Berrill \\
    Technical University of Berlin\\
    Berlin, Germany \\
  \And
    Felix Creutzig \\
    Technical University of Berlin\\
    Berlin, Germany \\
}

\begin{document}

\maketitle
\vspace*{-0.6cm}

\begin{abstract}
Understanding why travel behavior differs between residents of urban centers and suburbs is key to sustainable urban planning. Especially in light of rapid urban growth, identifying housing locations that minimize travel demand and induced CO\textsubscript{2} emissions is crucial to mitigate climate change. While the built environment plays an important role, the precise impact on travel behavior is obfuscated by residential self-selection.
To address this issue, we propose a double machine learning approach to obtain unbiased, spatially-explicit estimates of the effect of the built environment on travel-related CO\textsubscript{2} emissions for each neighborhood by controlling for residential self-selection. We examine how socio-demographics and travel-related attitudes moderate the effect and how it decomposes across the 5Ds of the built environment. Based on a case study for Berlin and the travel diaries of 32,000 residents, we find that the built environment causes household travel-related CO\textsubscript{2} emissions to differ by a factor of almost two between central and suburban neighborhoods in Berlin. To highlight the practical importance for urban climate mitigation, we evaluate current plans for 64,000 new residential units in terms of total induced transport CO\textsubscript{2} emissions. Our findings underscore the significance of spatially differentiated compact development to decarbonize the transport sector.

\end{abstract}

\section{Introduction}

The built environment plays a critical role in facilitating or hindering the decarbonization of urban transport \citep{IPCC_2022_WGIII_Ch_10}. The location and compactness of new residential development shape and constrain residents' future travel behavior. As such, they limit the mitigation potential of mobility-related lifestyle changes and potentially lock in transport-related emissions for decades \citep{creutzig2016urban}. Urban sprawl and suburban development, in particular, have been widely criticized for increasing car dependence and travel demand, hindering the decarbonization of urban transport \citep{dieleman2004compact}. Many studies have observed that residents of urban (central, higher-density, mixed-use) neighborhoods tend to drive less and instead walk \citep{van2012relationship, mccormack2011search, barnett2017built, cerin2017neighbourhood}, bike \citep{van2012relationship}, and use public transit more \citep{aston2021exploring, ibraeva2020transit} than residents of suburban (non-central, lower-density, single-use residential) neighborhoods. Yet, it is ambiguous to what extent these differences can be attributed to the built environment itself, as opposed to pre-existing differences in travel preferences between residents of urban and suburban neighborhoods. These differences are a result of a process known as residential self-selection \citep{caoExaminingImpactsResidential2009}, in which people choose their place of residence based on, among other things, locally available transport options matching their pre-existing travel preferences. Failing to account for residential self-selection and the resulting differences in travel preferences among residents of different neighborhoods can lead to falsely attributing observed differences in travel behavior to the built environment alone, overestimating its impact and drawing biased conclusions \citep{guanRoleResidentialSelfselection2020}.

Two strands in literature aim to uncover the built environment’s influence on travel behavior, one focusing on the nonlinearity of the relationship using machine learning methods, in particular decision tree ensembles \citep{aghaabbasi2023machine, cao2023using}, the other focusing on disentangling the influence from the confounding effect of residential self-selection using statistical methods such as statistical control, propensity score matching, and sample selection \citep{mokhtarian2008examining}. We aim to bring both approaches together.

% By using the causal inference method double machine learning (DML) \cite{chernozhukov2018double}, we leverage the flexibility of gradient boosting regression trees in modeling non-parametric, high-dimensional relationships while being able to control for confounding factors using Neyman-orthogonal scores and cross-fitting \citep{chernozhukov2017double}.
% Generally, DML is a framework to estimate heterogeneous treatment effects from observational data for multiple, continuous treatments while controlling for non-parametric confounding effects.
% Confounding factors are often identified from a directed acyclic graph (DAG).

We explore the potential of the causal inference method \textit{double machine learning} (DML) \cite{chernozhukov2018double} to examine the non-linear effect of the built environment on travel behavior from mobility survey data while accounting for confounding factors.
% We explore the potential of DML to examine the effect of the built environment on travel behavior from mobility survey data.
We assess the built environment’s impact across multiple, continuous dimensions, specifically the 5Ds \citep{ewing2010travel}, density, diversity, design, destination accessibility, and distance from transit, which all are presumed to have independent effects on travel behavior \citep{guanRoleResidentialSelfselection2020}. We control for the non-linear influence of socio-demographics and travel-related attitudes that are responsible for residential self-selection \citep{mokhtarian2008examining}.
% identify confounding factors from a causal directed acyclic graph (DAG) based on previous work and 
We examine the impact on travel behavior in terms of travel-related emissions to account for changes in both, travel distances and mode choice, allowing for a comprehensive assessment of the climate-related impacts of the built environment. We demonstrate our approach using the city of Berlin, Germany, as a case study and discuss local implications for low-carbon urban planning.
Our research questions are the following:
\begin{enumerate}
    \itemsep0em 
    \item What is the isolated effect of the built environment on travel behavior and induced CO\textsubscript{2} emissions when accounting for residential self-selection for each neighborhood in Berlin?
    \item How does the effect decompose into the 5Ds and how is it moderated by socio-demographics and travel-related attitudes?
    \item What are the induced transport CO\textsubscript{2} emissions of currently planned housing projects in Berlin and how can the spatial allocation be improved to most effectively reduce emissions?
\end{enumerate}

\section{Methods}

% Our goal is to obtain an unbiased estimate of the travel-related emissions for each neighborhood that can solely be attributed to the built environment. Our starting point is observed differences in travel-related emissions according to surveyed trip data. However, these differences are biased estimates of the true impact of the built environment because they do not account for individuals' pre-existing travel preferences that influence their residential choices (self-selection). Here, we use double machine learning to control for the confounding effects of residential self-selection by accounting for varying socio-demographics and attitudes in order to estimate the isolated effect of the built environment.

\subsection{Preprocessing}

\begin{figure}[!htbp]
    \centerline{\includegraphics[width=410pt]{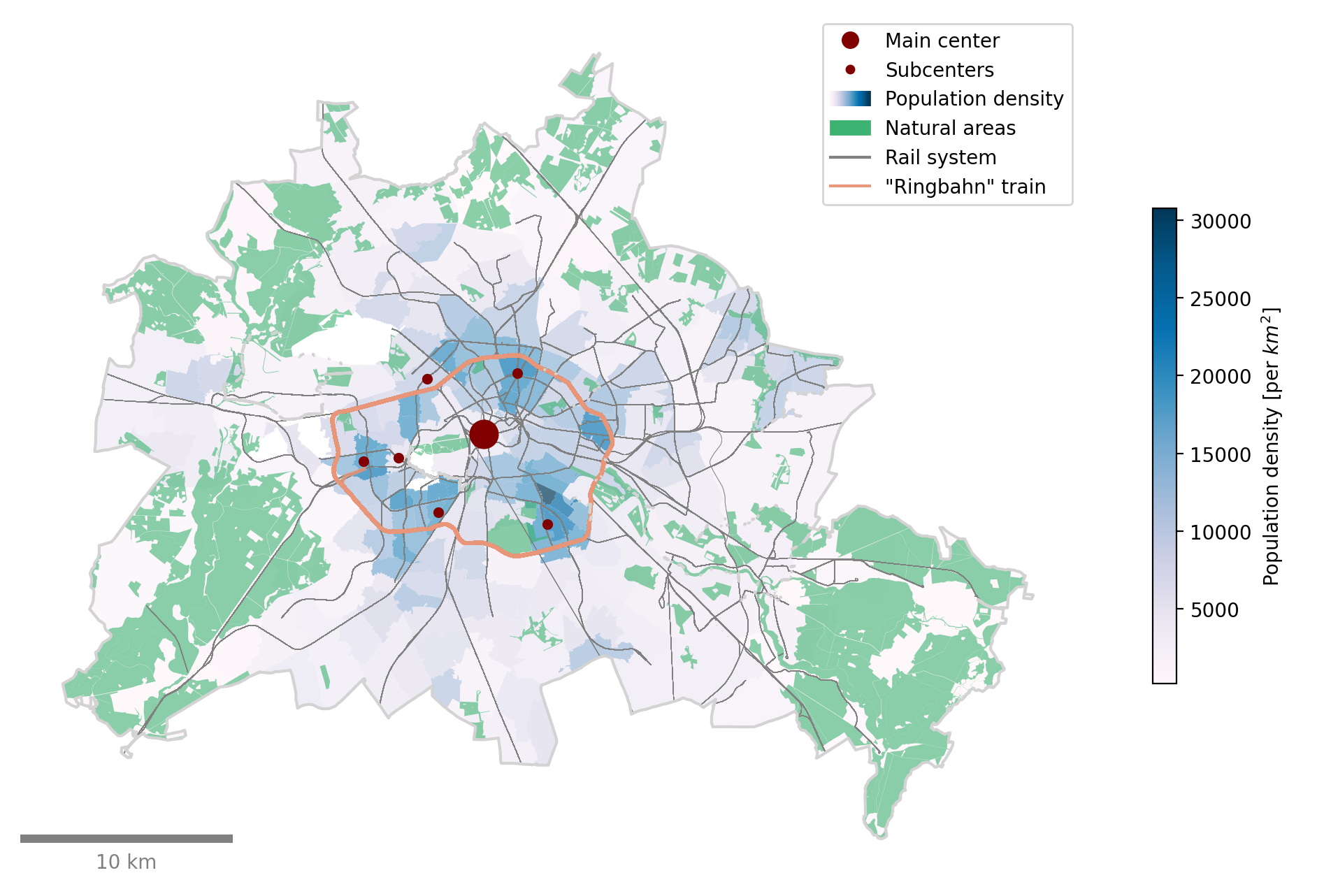}}
\figcaption{Built environment of Berlin, Germany.}{
The center and subcenters, based on points of interest density, are indicated as dark red circles.
Population density of neighborhoods is color coded in blue.
Natural areas according to Berlin land use data \citep{OffeneDatenBerlin} are marked in green.
The public transportation rail network is drawn in gray, with the exception of the so-called "Ringbahn", a commuter rail line that circles central Berlin, which is highlighted in orange.
We consider neighborhoods that are located outside of the "Ringbahn" and not within walking distance to be suburban.
}
\label{fig:study-area}
\end{figure}

\para{Data sources}. Representative travel behavior per neighborhood is obtained from travel diaries of the German national mobility survey "Mobilität in Städten – SrV 2018" \citep{hubrich2019methodenbericht}, which includes more than 100,000 trips for Berlin.
% The survey also includes socio-demographic characteristics such as age, income, education, and household size, as well as travel preferences such as car ownership, driving license, transit subscription, and bicycle ownership. Complementarily, we use the official vote shares for the Green party in the last regional election as a neighborhood level proxy for attitudes towards sustainable mobility and environmental issues, as the election debates were heavily characterized by the car vs. bike infrastructure trade-off.
To describe the built environment, we use only publicly available data from OpenStreetMap \citep{OpenStreetMap} on the street network and points of interest, from the public transit operator VBB on transit accessibility (GTFS data) \citep{gmbhGTFSDataBerlinBrandenburg} and from Berlin’s open data portal \citep{OffeneDatenBerlin} on population density, land-use, street greenery, street space allocation, and election results. See figure \ref{fig:study-area} for a visualization of the built environment of Berlin. Lastly, to convert mode-specific travel kilometers to emissions, we use emissions factors from the International Transport Forum (ITF) \citep{cazzola2020good} (see table \ref{table:emission-factors}).

\para{Unit of analysis.} We approximate the representative average travel behavior per neighborhood by averaging the surveyed and preprocessed travel behavior of all households. We assume that neighborhoods, in our case zip code areas, are sufficiently homogeneous in terms of their built environment to be able to detect a consistent impact on the travel behavior of its residents. This approach has the benefit of reducing noise related to sampling from a wide range of individual daily travel patterns and only examining to what degree the average travel behavior differs between neighborhoods.

% \para{Built environment characteristics.} We guide our characterization of the built environment by the 5Ds, Density, Diversity, Design, Destination accessibility, and Distance from transit, which all are presumed to have independent effects on travel behavior \citep{guanRoleResidentialSelfselection2020}. We refer to previous work and engineer one or more features for each dimension. To prevent density characteristics being biased by large green areas in otherwise densely populated neighborhoods, we only consider the built up area for any density calculation. A brief overview of the resulting urban form features is depicted in table \ref{table:built-env}.

\subsection{Causal inference}

\begin{figure}[!htbp]
    \centerline{\includegraphics[width=250pt]{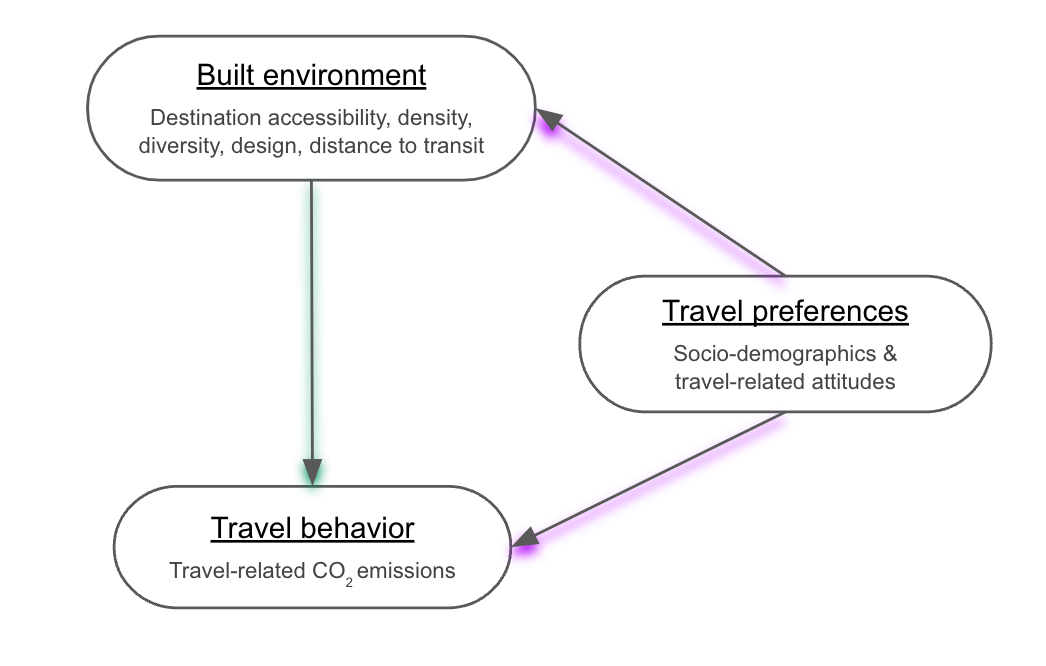}}
\figcaption{Directed acyclic graph (DAG).}{Travel preferences confound the effect of the built environment on travel behavior via residential self-selection. Operationalization of concepts as specified in the grey text.}
% {Directed acyclic graph (DAG) describing the causal relationship between the built environment, travel preferences, and travel behavior.
% (A) The built environment of a neighborhood, specifically the 5Ds, influence travel behavior of residents by modulating the mode share and the need to travel. (B) Pre-existing travel preferences influence the choice of residential location and thus the surrounding built environment (residential self-selection). (C) Pre-existing travel preferences influence travel behavior. 
% }
\label{fig:dag}
\end{figure}

\para{Confounding effects.} Travel preferences confound the influence of the built environment on travel behavior because of residential self-selection. The relationship is visualized as a directed acyclic graph (DAG) in figure \ref{fig:dag}. The confounding effect can be accounted for by controlling for socio-demographic traits and travel-related attitudes \citep{mokhtarian2008examining}. Here, we use information on age, income, education, and household size from the mobility survey to describe the socio-demographic composition of neighborhoods. Because travel-related attitudes are not directly captured in the survey, we use information on ownership of transport means as proxies, specifically car, bike, driving license, and transit subscription ownership.
% , hereafter referred to as transport means ownership,
% are the result of unobserved travel-related attitudes and therefore can serve as proxies.
While this is not a comprehensive characterization of travel-related attitudes, it has the advantage of capturing the temporal lag of travel-related attitudes through vehicle ownership, particularly car ownership \citep{van2014car}, that not only locks in travel behavior but also influences residential choice \citep{linBuiltEnvironmentTravel2017, scheiner2014residential}. Complementary to transport ownership, we capture general attitudes towards sustainable mobility and other environmental issues according to recent election results where the provision of sustainable transport infrastructure was in the public focus.
%and demand management solutions to reduce car usage such as city tolls and parking fees.
Refer to table \ref{table:confounding-factors} for a brief description of all attributes considered with respect to travel preferences.

% \input{confounding-factors}

% \para{Causal DAG.} We conceptualize the relationship between the built environment, travel preferences, and travel behavior as a directed acyclic graph (DAG), as shown in figure \ref{fig:dag}. It summarizes our assumptions about the data-generating process and facilitates confounder selection.
% We assume that individuals choose their residential location based on built environment characteristics that match their travel preferences (residential self-selection). Thus, their travel preferences influence their built environment choice. Since their travel preferences also influence their travel behavior, they confound the impact of the built environment on travel behavior.

\para{Treatment encoding.} 
To examine how the built environment of a specific neighborhood shapes travel behavior, we first determine how the local built environment is different from the city average. We characterize the difference along the 5Ds, density, diversity, design, destination accessibility, and distance from transit, for each of which we engineer at least one feature, guided by previous studies (see table \ref{table:built-env}).
To quantify how this difference impacts travel behavior, we define the treatment level as the difference between the neighborhood built environment and the city average built environment.
% In other words, we estimate a counterfactual world in which the built environment is homogeneous across the city and assess how this would change travel behavior. The change in travel behavior can be attributed to the built environment, whereas the remaining difference between neighborhood travel behaviors may be caused by residential self-selection or an incomprehensive characterization of the built environment.

% \input{built-env-features}

\para{Model selection.} Given continuous treatment along multiple treatment dimensions, we choose a CausalForest-based DML estimator to estimate the causal non-linear effect of the built environment on travel behavior.
DML \citep{chernozhukov2018double} consists of two stages: First, to account for the confounding effect, the outcome and treatment are being predicted from the controls using any appropriate machine learning model. Then, the treatment residuals are used to fit the outcome residuals yielding a debiased treatment effect estimate. The controls are also included to account for any moderating effects in explaining the heterogeneity of the treatment effect.
% the prediction residuals are used to fit a final model and estimate the causal, debiased effect of the treatment on the outcome. More specifically, the outcome residuals are predicted by the treatment residuals in combination with the other features to explain the heterogeneity. 
We use the DML open source implementation of the EconML package \citep{econml}. For the first stage, we use XGBoost \citep{chen2016xgboost} to model the non-parametric relationship between travel preferences and travel behavior and the built environment. We choose hyperparameters based on a random search using 5-fold cross validation, resulting in 1000 tree estimators, a tree depth of 6, and a learning rate of 0.01. For the final model, we use a CausalForest \citep{athey2019generalized} with 100 tree estimators.
%which also doesn’t impose any assumptions on the parametric form of the heterogeneous treatment effect.
% and offers confidence intervals via the Bootstrap-of-Little-Bags technique. 

\para{Effect heterogeneity \& composition.} To analyze the heterogeneity of the built environment’s effect on travel behavior, we calculate SHAP values \citep{lundberg2017unified} for the constant marginal effect estimation of the final stage model. This allows us to examine the moderating influence of confounding socio-demographic traits and travel-related attitudes on the effect. We are interested in the subgroups of households for which the built environment may have a particularly large or small effect. To avoid reporting of spurious correlations that the model picked up, we repeat the effect estimation 10 times, for all 5D dimensions combined and each individually, and only report moderators with a clear and consistent influence across all iterations. 
Further, we decompose the total effect of the built environment into the 5D dimensions and discuss the main drivers for low-carbon urban development. To make the 5D dimensions comparable in terms of their contribution to the overall marginal effect, we standardize all built environment characteristics and ensure consistent direction of feature values.
 % by removing the mean, scaling to unit variance

\subsection{Urban planning case study.}

To highlight the practical importance for urban climate mitigation, we evaluate current plans for 64,000 new residential units in Berlin in terms of total induced transport emissions. We compare the induced emissions of three alternative densification scenarios: (1) transit-oriented development focused on neighborhoods whose residents have an average walk time of less than 7 minutes to the nearest rail station, (2) transit-oriented development focused on centrally located neighborhoods, specifically neighborhoods connected by or located within the commuter rail line that circles central Berlin, the so-called “Ringbahn”, and (3) densification of low-emission neighborhoods where the built environment has the largest reducing impact on emissions according to our estimations. We assume that the housing units will be evenly distributed among the targeted neighborhoods for all three scenarios. We characterize the potential emissions savings from each strategy compared to the average household emissions. We assume that the built environment remains unchanged, i.e., we do not consider potential changes related to the new residential developments.

%%%%%%%%%%%%%%%%%%%%%%%%%%%%%%%%%%%%%%%%%%%%%%%%%%%%%%%%%%%%%%%%%%%%%%%%%%%%%%%%%%%%%%%%%%%%%%%%
%%%%%%%%%%%%%%%%%%%%%%%%%%%%%%%%%%%%%%%%%%%%%%%%%%%%%%%%%%%%%%%%%%%%%%%%%%%%%%%%%%%%%%%%%%%%%%%%
%%%%%%%%%%%%%%%%%%%%%%%%%%%%%%%%%%%%%%%%%%%%%%%%%%%%%%%%%%%%%%%%%%%%%%%%%%%%%%%%%%%%%%%%%%%%%%%%

\section{Results}

\begin{figure}[!htbp]
\begin{floatrow}
\ffigbox{%
\begin{adjustwidth}{-0.5in}{0in}
  \includegraphics[width=250pt]{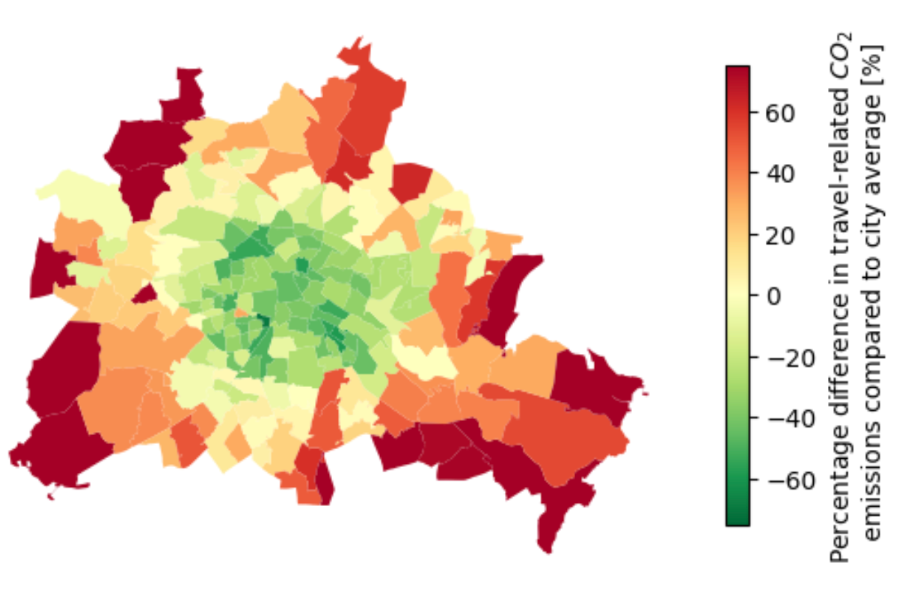}
  \vspace{-0.7cm}
\end{adjustwidth}
}{%
\figcaption{Spatial heterogeneity of the built environment's effect on travel-related CO\textsubscript{2} emissions.}{}
\label{fig:spatial-effect}
}
\capbtabbox{%
  \input{effect-decomposition}

}{%
\tablecaption{Decomposition of the built environment’s overall constant marginal treatment effect}{}
\label{table:effect-decomposition}
}
\end{floatrow}
\end{figure}

\para{Isolated effect.} The built environment has a considerable effect on travel behavior and related emissions after controlling for residential self-selection. In the center of Berlin, the built environment facilitates a 40\% decrease and in the outskirts a 50\% increase in travel-related emissions compared to the city average (see figure \ref{fig:spatial-effect}). Overall, half of the observed differences in neighborhood emissions are due to the built environment according to our CausalForest-based DML estimator. The remaining residuals may partly be due to residential self-selection and an insufficient characterization of the built environment. Overall our features explain the variation of neighborhood emissions with a coefficient of determination (R\textsuperscript{2}) of $0.85$, suggesting a good characterization of travel preferences and the built environment. The nuisance scores of the first stage DML models indicate a high degree of confounding, with the R\textsuperscript{2} being $0.76$ and $0.56$ for fitting the target and treatment, respectively.
% In Berlin, we find that emissions in the center are around 40\% lower than average household travel-related emissions due to the built environment, while they are more than 50\% larger in the outskirts of the city (see figure \ref{fig:spatial-effect}).
% Overall, half of the observed differences in neighborhood emissions can be attributed to the built environment; 

% \begin{figure}[!htbp]
%     \centerline{\includegraphics[width=410pt]{spatial-effect.png}}
% \figcaption{Built environment significantly impacts travel behavior and related emissions.}{
% Left: Relative impact on travel-related emissions attributable to the built environment of each neighborhood compared to city average. Right: Unexplained residual between observed emission differences from city average and effect estimate of the built environment.
% % Presumably caused by insufficient characterization of the built environment and residential self-selection (RSS).
% }
% \label{fig:spatial-effect}
% \end{figure}

\begin{figure}[!htbp]
    \centerline{\includegraphics[width=400pt]{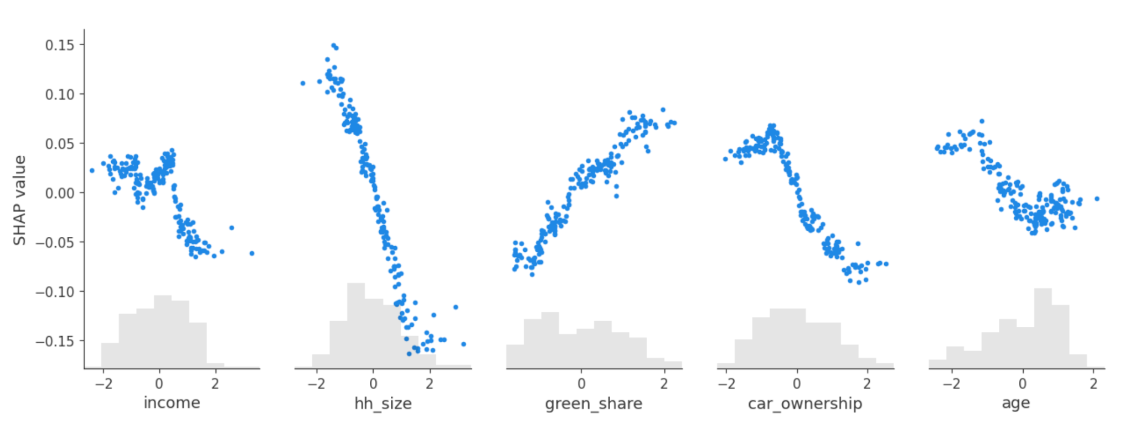}}
\figcaption{Household size, income, age, car ownership, and environmental attitudes consistently moderate treatment effect.}{
SHAP dependence plot for the final stage model predicting treatment heterogeneity. Selection of confounders that have a consistent moderating effect on travel-related emissions. Confounding variables have been standardized to facilitate comparison. A SHAP value below 0 corresponds to an increased effect of the built environment for such households, whereas a SHAP value above zero corresponds to a decreased effect. Consistent zero SHAP values for all household characteristics would imply that there is no effect heterogeneity. 
}
\label{fig:heterogeneity}
\end{figure}

\para{Effect heterogeneity.} The estimated effect of the built environment is moderated by socio-demographics and travel-related attitudes. The household size, income, age and car ownership are positively associated with the impact of distance to the city center on travel-related emissions meaning that the impact of the built environment may be larger for households that are relatively large, old, high-income, or highly car-owning (see figure \ref{fig:heterogeneity}). On the other hand, environmentally friendly attitudes are associated with a lower effect of the built environment. The other confounding factors do not exhibit a consistent impact on the heterogeneity across multiple iterations with different seeds and are thus excluded from our results.

\para{Effect composition.} From the 5Ds, destination accessibility has the largest effect on travel-related emissions. 73.7\% of the built environment’s influence is determined by the distance to the center, to the subcenter and the local neighborhood job accessibility (see table \ref{table:effect-decomposition}). The second most important D is density, with the adjusted population density being responsible for 15.2\% of the total effect. While design and distance from transit have a small effect, 6.4\% and 4.3\% respectively, diversity in terms of the mixed-use land share has no meaningful impact on travel-related emissions according to our approach.

\begin{figure}[!htbp]
    \centerline{\includegraphics[width=410pt]{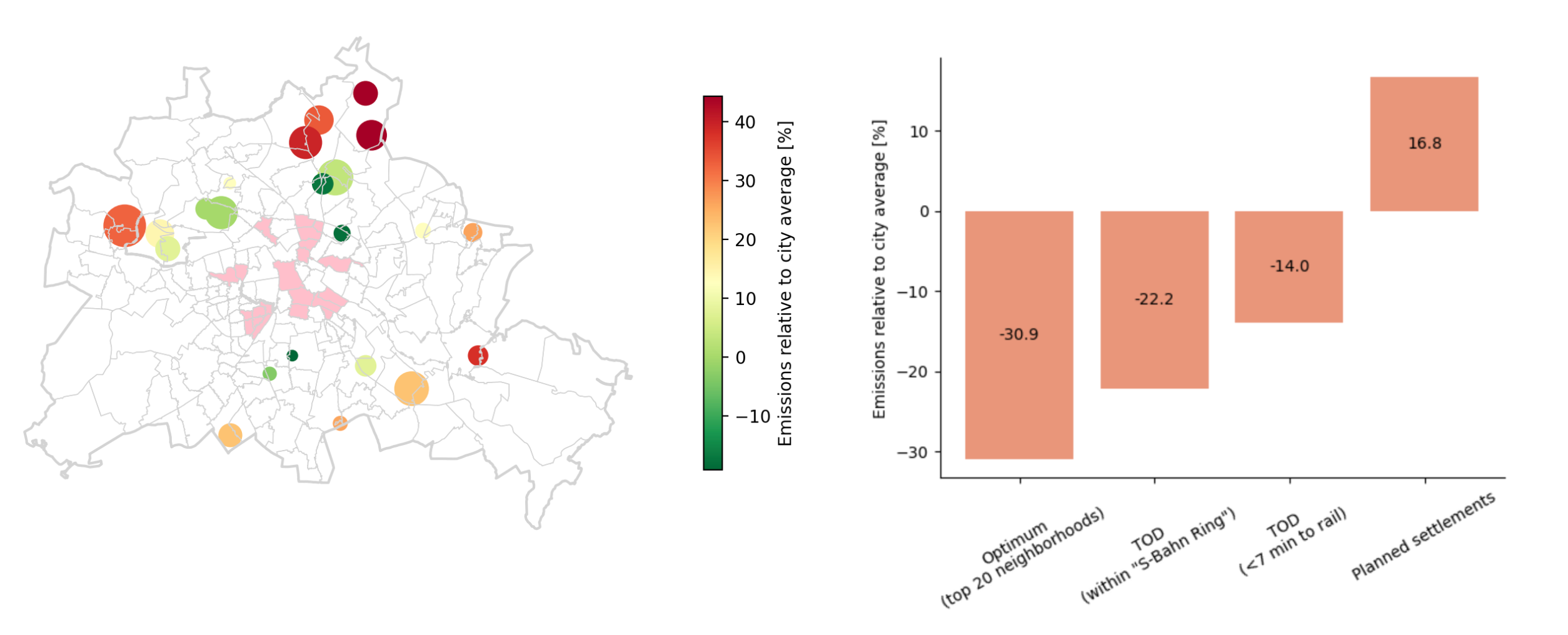}}
\figcaption{Induced transport CO\textsubscript{2} emissions of planned settlements 70\% above the theoretical optimum.}{
Left: New settlement locations are marked by circles, with color representing future residents' expected average emissions and size indicating planned housing units. Neighborhoods with the largest reducing effect on emissions, targeted by the “Optimum” policy, are marked in light pink.
Right: Comparison of different residential planning strategies, including transit-oriented development (TOD) and a densification according to the model (“Optimum”), in terms of average household travel-related emissions relative to the city average.
% The two transit-oriented development (TOD) strategies focus residential development in neighborhoods with good transit accessibility: either less than 7 minutes walking time to rail-based transit or within the so-called “Ringbahn”, a commuter rail that encircles the center.
% “Planned settlements” refers to newly planned urban quarters with approximately 64,000 housing units scheduled for completion by 2030.
}
\label{fig:scenarios}
\end{figure}

\para{Urban planning case study.} Due to location-dependent built environment effects, the planned settlements in Berlin are expected to increase average household travel-related emissions by 16.8\% above the city average (see figure \ref{fig:scenarios}). Alternative residential development strategies that focus on densifying the center and transit-oriented development may lead to significantly lower emissions. We estimate that emissions can be reduced by up to 30.9\% if residential planning prioritizes the 20 neighborhoods with the most sustainable built environment according to the model. If all neighborhoods with a good transit accessibility are targeted, specifically within 7 minutes of walking to the nearest rail station, emissions of future residents are expected to be 14\% below the city average.
% If the walking time threshold is lower, emissions are further reduced along a sigmoidal curve.

\section{Discussion \& Conclusion}

With the need to reduce transportation emissions and curb global boiling, growing cities face the urgent question of where to locate new residents to minimize travel demand and related emissions. Thus, having a spatially explicit estimate of the built environment's impact on transport emissions at potential new residential development planning sites is indispensable for evidence-based low-carbon urban planning. 

In Berlin, the impact of the built environment on travel-related emissions is tremendous. Household emissions differ by a factor of two between the city center and the outskirts because of the built environment. This disparity is mainly due to the different accessibility of destinations (74\%).
% followed by differences in population density (15\%).
We find that the effect of the built environment is largest for households that are relatively large, old, high-income, or car-owning. According to our calculations, the induced emissions of the currently planned 66,000 residential units are 70\% above the optimum. Alternative compact or transit-oriented development strategies would lead to significantly lower emissions.

While we believe that the overall effect magnitude is robust, we want to emphasize that the precise estimates of induced emissions are subject to uncertainty for three main reasons. First, causal inference assumptions are partially violated, most importantly we do not account for spatial interference of treatments and spatial confounding effects \citep{reich2021review}. Second, travel-related attitudes are likely affected to some degree by the built environment and past travel behavior and thus interdependent instead of being an exogenous predisposition \cite{naessResidentialSelfSelection2009, scheinerTransportCostsSeen2018}. Third, by aggregating travel behavior and built environment characteristics for each zip code area, we mask heterogeneous distributions and nuances in the disaggregated data, potentially leading to a biased estimate (MAUP effect \citep{fotheringham1991modifiable, zhang2005metrics}).

% In conclusion, we believe that double machine learning offers cities a cost-effective tool that greatly facilitates the estimation of travel demand induced by the built environment and, although subject to some uncertainty, can establish a starting point for evidence-based sustainable urban planning.
% In conclusion, we believe that double machine learning, although subject to some uncertainty, can greatly facilitate and scale the estimation of travel demand induced by the built environment and provide cities with a cost-effective tool to establish a starting point for evidence-based sustainable urban planning.

In conclusion, double machine learning has the potential to greatly facilitate and scale the estimation of travel demand induced by the built environment. Although the effect estimates are subject to some uncertainty, it can provide cities with a cost-effective tool to establish a starting point for evidence-based sustainable urban planning.
% With its high spatial resolution, it can further improve the spatial differentiation of  compact development.

\begin{ack}
This work received funding from the CircEUlar project of the European Union’s Horizon Europe research and innovation program under grant agreement 101056810.
\end{ack}

\bibliography{bibliography}

\newpage
\section*{Appendix}
\FloatBarrier

\input{emission-factors}

\input{confounding-factors}

\input{built-env-features}

\end{document}

%% file: effect-decomposition.tex
% \begin{table}[!htbp]
\hskip0.3in
\centering
\renewcommand{\arraystretch}{1.3}
\footnotesize
\begin{tabular}{| p{1.55cm} | p{2.8cm} | p{1cm} |}
\hline
\textbf{5D} & \textbf{Feature name} & \textbf{Effect share} \\
\hline
\multirow{3}{*}{\shortstack[l]{Destination \\ accessibility}} & Distance to center & 51.2\% \\
 % \cline{2-3}
 & Distance to subcenter & 15.2\% \\
 % \cline{2-3}
 & POI density index & 11.1\% \\
\hline
Density & Population density & 11.4\% \\
\hline
Diversity & Land use & 0.3\% \\
\hline
\multirow{2}{*}{Design} & Car-friendliness index & - \\
% \cline{2-3}
 & Walkability index & 6.4\% \\
\hline
Distance to transit & Transit accessibility index & 4.3\% \\
\hline
\end{tabular}

% \tablecaption{Destination accessibility has the strongest impact on travel-related CO\textsubscript{2} emissions.}{
% Decomposition of the overall constant marginal treatment effect into the different dimensions of the built environment.
% }
% \end{table}

%% file: emission-factors.tex
\begin{table}[!htbp]
    \renewcommand{\arraystretch}{1.3}
    \small
    \centering
    \begin{tabular}{cc}
        \textbf{Mode} & \textbf{Emissions [g CO\textsubscript{2}/pkm]} \\
        \hline
        Car (ICE) & 162 \\
        Moped (ICE) & 70 \\
        Transit & 65 \\
        Bike & 20 \\
        Foot & 0 \\
    \end{tabular}
    \tablecaption{Emission factors of transport modes}{Central estimates of life-cycle greenhouse gas emissions of urban transport modes per person km according to the International Transport Forum (ITF) \citep{cazzola2020good}. Emissions factors are expressed CO\textsubscript{2} equivalents and have partially been aggregated to match transport modes considered in this study (e.g. bus \& metro \textrightarrow{} transit). Life-cycle emissions include a vehicle, fuel, and infrastructure component as well as operational services. ICE refers to internal combustion engine.}
    \label{table:emission-factors}
\end{table}

%% file: confounding-factors.tex
\begin{table}[!htbp]
\footnotesize
\renewcommand{\arraystretch}{1.3}
\begin{adjustwidth}{-0.7in}{0in}
\centering

\begin{tabular}{| l | l |  p{8.5cm} |}
\hline
\textbf{Category} & \textbf{Variable name} & \textbf{Description} \\
\hline
\multirow{4}{*}{Socio-demographics} & \texttt{income} & Average household income \\
% \cline{2-3}
 & \texttt{hh\_size} & Average number of persons living in a household \\
 & \texttt{age} & Average age of adult (>18 years) residents \\
 & \texttt{uni\_share} & Share of people older than 25 with university degree   \\
\hline
\multirow{7}{*}{Proxies for travel-related attitudes} & \texttt{car\_ownership} & Average number of private \& company cars per household \\
% \cline{2-3}
 & \texttt{bike\_ownership} & Average number of bicycles owned per person \\
 & \texttt{driving\_license} & Average share of adults (>18 years) with driving license \\
 & \texttt{transit\_subscription} & Average share of people with monthly transit subscription (incl. children and people with disabilities with free ride tickets) \\
 & \texttt{green\_share} & Electoral share of the Green party in constituencies intersecting the neighborhood in the last regional elections \\
\hline
\end{tabular}
\end{adjustwidth}

\tablecaption{Travel preferences}{Overview of all socio-demographic traits and proxies for travel-related attitudes considered in the study including a brief description. Children are excluded from the calculation of average age, education, and driver's license ownership to reduce the correlation between the variables and ensure that each variable describes a separate aspect.}
\label{table:confounding-factors}
\end{table}

%% file: built-env-features.tex
\begin{table}[!htbp]
\begin{adjustwidth}{-0.35in}{0in}
\centering
\renewcommand{\arraystretch}{1.3}
\footnotesize
\begin{tabular}{| l | l | p{7cm} |}
\hline
\textbf{5D’s of compact development} & \textbf{Feature name} & \textbf{Description} \\
\hline
\multirow{3}{*}{Destination accessibility} & Distance to center & Distance to neighborhood with highest POI density \\
 % \cline{2-3}
 & Distance to subcenter & Least distance to any of the 10 neighborhoods with highest POI density \\
 % \cline{2-3}
 & POI density index & Local POI density for offices, schools, kindergarten, and universities \\
\hline
Density & Population density & Population density of the built-up area \\
\hline
Diversity & Land use & Share of mixed-use areas \\
\hline
\multirow{2}{*}{Design} & Car-friendliness index & Provision of expressway kilometers per capita \\
% \cline{2-3}
 & Walkability index & Intersection density in the built-up area \\
\hline
Distance from transit & Transit accessibility index & Gravity model-based index describing the average spatio-temporal transit accessibility of a neighborhood \\
\hline
\end{tabular}
\end{adjustwidth}

\tablecaption{Built environment characteristics}{Overview of all built environment characteristics considered in the study including a brief description and mapping to the corresponding D of compact development. The street network and points of interest (POIs) are extracted from OpenStreetMap. Density measures are calculated using the built up area only. 
Indices are calculated using sklearn's \texttt{StandardScaler} \citep{scikit-learn}.
%to avoid biased by large green areas in otherwise densely populated neighborhoods, we only consider the built up area for any density calculation.
}
\label{table:built-env}
\end{table}